\documentstyle[icassp91,times,lingmacros]{article}

\title{\LARGE \bf{Evaluating Competing Agent Strategies for a Voice Email Agent}}
\name{Marilyn Walker, Donald Hindle, Jeanne Fromer, Giuseppe Di Fabbrizio, Craig Mestel}

\address{AT\&T Labs -- Research \\
180 Park Ave, Florham Park, NJ 07932, USA \\
email: {\tt \{walker,hindle,pino\}@research.att.com}}

\newcommand{\beq}[1]{\begin{equation}\label{#1}}
\newcommand{\eeq}{\end{equation}}

\begin{document}
\begin{sloppypar}
\maketitle
\input{psfig}
\bibliographystyle{plain}
\begin{abstract}

This paper reports experimental results comparing a mixed-initiative
to a system-initiative dialog strategy in the context of a personal
voice email agent. To independently test the effects of dialog
strategy and user expertise, users interact with either the
system-initiative or the mixed-initiative agent to perform three
successive tasks which are identical for both agents. We report
performance comparisons across agent strategies as well as over
tasks. This evaluation utilizes and tests the PARADISE evaluation
framework, and discusses the performance function derivable from the
experimental data.

\end{abstract}

\section{Introduction}

Recent advances in spoken language processing, text-to-speech,
dialog modeling, and computational power have led to the development
of spoken dialog agents for many types of information services.
Much work to date has focused on the development and evaluation of the
component technologies and on the non-trivial task of making them work
together.  This has led to a number of fully integrated systems that
can be experimentally evaluated with real users carrying out
representative tasks.  However, continued progress in understanding what design
choices make the most effective systems requires a general spoken
dialog evaluation framework that supports the quantitative
comparison of different agent designs in the context of actual use.
For example consider the two alternative dialog strategies for a
voice email agent, exemplified by the dialogs in D\ex{1} and D\ex{2}
(A is the agent and U is the user):

\enumsentence[(D1)]{
A: Hi, Elvis here.  You have 5 new and 0 unread messages in your
inbox. Say
Repeat to repeat this  message, or say
Read, Summarize, or I'm done here. \\
U: Read. \\
A: Select by Content or Position? \\
U: Content. \\
A: Select by Sender or Subject? \\
U: Sender \\
A: Which Sender?\\
U: Kim.\\
A: The message from Kim is about Meeting Tomorrow. The meeting tomorrow is at 10:30 in 2D-516.}

\enumsentence[(D2)]{
A: Hi, Elvis here. I've got your mail. \\
U: Read me my messages from Kim. \\
A: The message from Kim is about Meeting Tomorrow. The meeting tomorrow is at 10:30 in 2D-516.
}

Both of these agents support the same task functionality for voice
access to email, and are implemented in a system called ELVIS (EmaiL
Voice Interactive System). The Elvis agent in D1 is based on a
system-initiative dialog strategy, whereas the Elvis agent in D2 is based
on a mixed-initiative dialog strategy. The system-initiative
strategy acquires information in small increments and constrains the
user to single word utterances whose content is explicitly prompted
for. The mixed-initiative strategy lets the user control the dialog,
doesn't provide information unless the user asks for it, and allows
the user random access to all the application functionality with
utterances that combine a set of information requirements.

It may seem obvious that the mixed-initiative strategy in D2 is
preferable to the system-initiative strategy in D1. Previous work has
emphasized the utility of mixed-initiative dialog strategies in
advice-giving and diagnostic dialog domains
\cite{WW90,SmithHipp}. However, other work suggests that the
performance of the system-initiative agent may be superior
\cite{Marcus96,DanieliGerbino95,PRBO96}. One reason for this is the
less than perfect performance of current speech recognizers.  The
mixed-initiative strategy requires more complex grammars, possibly
resulting in higher automatic speech recognition (ASR) error
rates. This in turn may lead to a higher overall task error rate, or
extremely long repair subdialogs.  A second potential problem is
that the mixed-initiative strategy may require users to learn what the
system can understand, since the system does not explicitly prompt
them with valid vocabulary.

This suggests that the mixed-initiative strategy may be more suitable
for experienced users. However, spoken dialog agents have rarely been
evaluated in the context of repeated use by a single user
\cite{SWP92}, as would be expected in the case of an email agent.  It
is likely that (1) users become more expert over time; and (2) in the
future systems will adapt and learn. Thus it is important to evaluate
changes in performance over repeated user sessions.  We hypothesize
that the more experience a user has with the system, the better the
mixed-initiative strategy will perform.

This paper describes the implementation of these two dialog
strategies in an agent for accessing email by phone. We present the
results of an experiment in which users perform a series of tasks by
interacting with an email agent using one of the dialog strategies.
We also describe how our experimental results can be framed in the
PARADISE framework for evaluating dialog agents.

\section {System Design}
\label{sd-sec}

In order to determine the basic application requirements for email
access by telephone, we conducted a Wizard of Oz study.  The Wizard
simulated an email agent interacting with six users who were
instructed to access their email over the phone at least twice over a
four-hour period.  In order to acquire a basic task model for email
access over the phone, the Wizard was not restricted in any way, and
users were free to use any strategy to access their mail.  The
study resulted in 15 dialogs, consisting of approximately 1200
utterances, which were transcribed and analyzed for key email access
functions.

We categorized email access functions into general categories based on
the underlying application, as well as language-based requirements,
such as the ability to use referring expressions to refer to messages
in context (as {\it them, it, that}), or by their properties such as
the sender or the subject of the message.  Table 1 summarizes the
functions used most frequently in our Wizard of Oz study; these
frequencies were used to prioritize the design of the email application
module.

\begin{small}
\begin{table}[htb]
\begin{center}
\begin{tabular}{|l|c|}
 \hline
Function& N  \\
 \hline \hline
Summarization& 20 \\
\hline
Reference& 101\\
\hline
Folder Action& 10 \\
\hline
Read Message& 67 \\
\hline
Message Field& 5 \\
\hline
Repeat& 4 \\
\hline
Clarification by User& 24 \\
\hline
Clarification by Agent& 13\\
\hline
Search for a message& 8 \\
\hline
Help& 3 \\
\hline
\end{tabular}
\caption{Frequency of Email Functions from Wizard-Of-Oz Study over all dialogs}
\end{center}
\label{freq-fig}
\end{table}
\end{small}

From this exploratory study we concluded that the email agent should
minimally support: (1) reading the body of a message and the header
information; (2) summarization of the contents of an email folder by
content-related attributes, like sender or subject; (3) access to
individual messages by content fields such as sender and subject; (4)
requests for cancellation and repetition by the user and for
clarifying help from the system.

We implemented both the system-initiative and the mixed initiative
versions of the email agent within a general-purpose platform for
voice dialog agents, which combines ASR, text-to-speech (TTS), a
phone interface, an email access application module, and modules for specifying the
dialog manager and the application grammars \cite{Kammetal97}. The
email application demands several advanced capabilities from these
component technologies. First, ASR must support barge-in, so that the
user can interrupt the agent when it is reading a long email
message. Second, the agent must use TTS due to the dynamic and
unpredictable nature of email messages; prerecorded prompts are not
sufficient for email access. Third, the grammar module must support
dynamic grammar loading because the ASR vocabulary must change to
support selection of email messages by content fields such as sender
and subject.

The application module is implemented as a separate module called the
Email Application Interface (EMAI). EMAI is compatible with several
e-mail server protocols, e.g.  POP3, IMAP4 and SMTP, and is based on
the standard known as RFC822. It also decodes the MIME standard for
attachments (Text, Graphics, Application specific, Video, Sound), and
provides the information back to the application.

EMAI provides a set of basic email access functions to the dialog
manager. One feature is {\bf access} to message attributes such as
subject, author, body, unique message ID, and any attachments. A
second set of features supports {\bf selection} of messages by message
attributes such as subject or sender, or by positional attributes such
as previous, next, or last. Third, sets of messages can be {\bf
sorted} by message attributes such as the Author's Reply Address,
Date, Subject, Status, Length, or Priority. Fourth, messages and
message folders can have their status information {\bf updated}. For
example, after the agent reads a message to a user the status may be
changed from {\it new} to {\it read}, or after the user says `delete
it', the status is modified from {\it new} to {\it deleted}. Fifth,
the application module {\bf preprocesses} the message body for TTS to
modify material not easily realized in speech \cite{Beutnagel93}.
Finally, the recursive structure of dialog requires EMAI to manage a
stack of folders reflecting the dialog structure.  The main
advantage of EMAI to the dialog manager is that it abstracts from
the implementation of the application system, so that the same
application module can be used with different underlying mail systems.
Both versions of the email agent have identical task functionality.

The dialog manager uses a state machine to implement both the
system-initiative and the mixed-initiative dialog strategy. Each
state includes parameter specifications for: (1) an initial prompt,
which the agent says upon entering the state (this may include a
response to the user's current request); (2) whether barge-in is
enabled; (3) a help prompt which the agent says if the user says {\it
help}; (4) multiple rejection prompts which the agent  says if ASR
rejects the user's utterance; (5) multiple timeout prompts which the
agent  produces if the user doesn't say anything; and (6) a grammar
specifying what the user can say. Transitions between states are
driven by the semantic interpretation of user's utterances.

There are two main differences between the mixed-initiative and
system-initiative dialog strategies. First, the mixed-initiative
agent does not volunteer information in its initial prompt, or
explicitly tell the user what to say. This information is obtainable
at the user's initiative by waiting for the timeout prompt to play or
by saying {\it Help}. Secondly, as shown in D1 and D2, the
system-initiative agent severely constrains the recognition grammar
available at any point, and prompts the user for each increment of
information needed to carry out an application function.  In contrast,
the mixed-initiative agent allows most responses at all times, and
allows the user to speak in full sentences which specify multiple
arguments to application functions simultaneously.  In terms of the
state machine, the mixed-initiative version's one main state
corresponds to 8 system-initiative states, reflecting the fact that
the system-initiative version is more restrictive and has more
constrained ASR grammars.

\section{Experimental Design}

The experiment required users, randomly assigned to one agent or the
other, to complete three tasks involving telephone access to email, in
three separate conversations with the agent. All of the users
regularly used computers in the course of their everyday work and were
familiar with email.  Instructions were given on three web pages, one
for each experimental task. Each web page consisted of a brief general
description of Elvis, a list of hints for using Elvis, a task
description, and information on calling Elvis.  Each page also
contained a form for specifying information acquired from the agent
during the dialog, and a survey, to be filled out after task
completion, designed to probe the users' satisfaction with the
system. Subjects read the instructions in their offices before calling
Elvis from their phone.

Each user performed three tasks in sequence, and each task consisted
of two subtasks.  Six users were assigned to the mixed-initiative
agent and 6 users to the system-initiative agent. Thus the experiment
resulted in 36 dialogs representing 72 attempted subtasks.  To be
consistent with the PARADISE evaluation framework \cite{Walkeretal97},
each subtask was represented by a scenario where the agent and the
user had to exchange information about criteria for selecting messages
and information within the message body. For example, in one scenario
the user is expecting email from Kim about a meeting and needs to find
out the time and place of that meeting (as in Dialog D1 and
D2). This scenario is represented in terms of the attribute value
matrix (AVM) in Table 2.

{\small 
\begin{table}[htb]
\begin{center}
\begin{tabular}{|l|l|} \hline
attribute               &  value       \\ \hline
Selection Criteria          &Kim $\vee$ Meeting\\
Email.att1              &10:30       \\
Email.att2              &       2D516   \\ \hline
\end{tabular}
\label{key}
\caption{Attribute Value Matrix: Email Scenario Key for Dialogs 1 and 2 }
\end{center}
\end{table}
}

The task scenarios that the subjects were given were as follows, where
scenarios 1.1 and 1.2 were done in the same conversation with Elvis, similarly
for 2.1 and 2.2, and 3.1. and 3.2.

\begin{itemize}

\item TASK 1.1: You are working at home in the morning and plan to go
directly to a meeting when you go into work.  Kim said she would
send you a message telling you where and when the meeting is.  Find
out {\bf  the Meeting Time} and {\bf the Meeting Place}.

\item TASK 1.2: The second task involves finding information in a
different message. Yesterday evening, you had told Lee you might want
to call him this morning.  Lee said he would send you a message
telling you where to reach him.  Find out {\bf Lee's Phone Number}.

\item TASK 2.1: When you got into work, you went directly 
to a meeting. Since some people were late, you've decided to
call Elvis to check your mail to see what other  meetings
may have been scheduled. Find out the day, place and time of any
scheduled meetings.

\item TASK 2.2: The second task involves finding information in a
different message. Find out if you need to call anyone. 
If so, find out the number to call.

\item 
 TASK 3.1: You are expecting a message telling you when
the Discourse Discussion Group can meet.  Find
out the {\bf place} and {\bf time} of the meeting.

\item TASK 3.2: The second task involves finding information in a
different message. Your secretary has taken a phone call for you
and left you a message. Find out  {\bf who called}  and  {\bf where} you can
reach them.

\end{itemize}

Successful completion of a scenario requires that all attribute-values
must be exchanged. The sender and subject attributes that are usable
as selection criteria are known by the user at the beginning of the
dialog, while the attributes to be extracted from the body of the
email message are acquired from the agent in the course of the
interaction.  The AVM representation for all six subtasks is similar
to Table 2. Note that the task's information-exchange requirement
represented in the AVM is independent of the agent strategy used to
accomplish the task.


Experimental results were collected by three means, and a series of
variables were extracted. First, all of the dialogs were
recorded. This allows utterance transcription and checking aspects of
the timing of the interaction, such as whether there were long delays
for agent responses, and whether users barged in on agent utterances
{\bf Barge In}. In addition, it was used to calculate the total time
of the interaction (the variable named {\bf Elapsed Time}).

Second, the system logged the agent's dialog behavior on the basis
of entering and exiting each state in the state transition table for
the dialog.  For each state, the system logged the number of timeout
prompts ({\bf Timeout Prompts}) , {\bf ASR Rejections}, and the times the user said {\it
Help} ({\bf Help Requests}). The
number of {\bf System Turns} and the number of {\bf User Turns} were calculated on
the basis of this data. In addition, the ASR result for the user's
utterance was logged. The recordings were used in combination with the
logged ASR result to calculate a concept accuracy measure for each
utterance.  Mean concept accuracy was then calculated over the whole
dialog and used as a {\bf Mean Recognition Score} for the dialog.

Third, users were required to fill out the web page forms after each
task specifying whether they had completed the task and the
information they had acquired from the agent ({\bf Task Success}),
e.g.  the values for Email.att1 and Email.att2 in Table 2. In
addition, users responded to a survey on their subjective evaluation
of their satisfaction with the agent's performance with the following
questions:
\begin{itemize}
\item Was Elvis  easy to understand in this conversation?  ({\bf TTS Performance}) 
\item In this conversation, did  Elvis  understand what you said? ({\bf ASR Performance}) 
\item In this conversation, was it easy to find the message you
wanted? ({\bf Task Ease}) 
\item Was the pace of interaction with Elvis appropriate in this
conversation? ({\bf Interaction Pace}) 
\item In this conversation, did you know what you could say at each
point of the dialog? ({\bf User Expertise}) 
\item  How often was Elvis sluggish and slow to reply to you in this
conversation? ({\bf System Response}) 
\item  Did Elvis  work the way you expected him to in this
conversation? ({\bf Expected Behavior}) 
\item In this conversation,  how did  Elvis's voice interface compare
to the touch-tone interface to voice mail? ({\bf Comparable Interface}) 
\item From your current experience with using Elvis to get your email,
do you think you'd  use Elvis  regularly to access your mail when you
are away from your desk? ({\bf Future Use}) 
\end{itemize}

Most question responses ranged over values such as ({\it almost
never, rarely, sometimes, often, almost always}), or an equivalent
range.  Each of these responses was mapped to an integer in 1
$\ldots$ 5.  Some questions had ({\it yes, no, maybe}) responses. Each
question emphasized the user's experience with the system in the
current conversation, with the hope that satisfaction measures would
indicate perceptions specific to each conversation, rather than
reflecting an overall evaluation of the system over the three tasks.
We calculated a {\bf Cumulative Satisfaction} score for each dialog by
summing the scores for each question.

\section{Experimental Results}
\label{eval-sec}

Our goal was to compare performance differences between the
mixed-initiative strategy and the system-initiative strategy, when the
task is held constant, over a sequence of three equivalent tasks in
which the users might be expected to learn and adapt to the system. We
hypothesized that the mixed-initiative strategy might result in lower
ASR performance, which could potentially reduce the benefits of user
initiative. In addition, we hypothesized that users might have more
trouble knowing what they could say to the mixed-initiative agent, but
that they would improve their knowledge over the sequence of
tasks. Thus we hypothesized that the system-initiative agent might be
superior for the first task, but that the mixed initiative agent would
have better performance by the third task.

Our experimental design consisted of three factors: strategy, task and
subject. Each of our result measures were analyzed using a three-way
ANOVA for these three factors. For each result we report F and p
values indicating the statistical significance of the results. Effects
that are significant as a function of strategy indicate differences
between the two strategies. Effects that are significant as a function
of task are potential indicators of learning. Effects that are
significant by subject may indicate problems individual subjects may
have with the system, or may reflect differences in subjects' attitude
to the use of spoken dialog interfaces.  We  discuss each of these
factors in turn.

We first calculated Task Success using the $\kappa$ (Kappa) statistic \cite{Walkeretal97}.

\[\kappa =  \frac {P(A) - P(E)}{1 - P(E)} \]

P(A) is the proportion of times that the AVMs for the actual set of
dialogs agree with the AVMs for the scenario keys, and P(E) is the
proportion of times that we would expect the AVMs for the dialogs
and the keys to agree by chance. Over all subjects, tasks and
strategies, P(E) = .50, P(A) is .95, and $\kappa$ is .9. Thus users completed
the assigned task in almost all cases.

Results of ANOVA by strategy, task and subject showed that there were
no significant differences for any factors for Task Success
(Kappa), Cumulative Satisfaction, or Elapsed Time to complete the
task. However there are differences in the individual satisfaction
measures, which we  discuss below. We believe the lack of an
effect for Elapsed Time reflects the fact that the dominant time
factor for the system is the email access application module, which was
constant across strategy.

{\bf Strategy Effects}: As we hypothesized, the Mean Recognition Score for the
system-initiative strategy (SI) was better than the Mean Recognition Score
for the mixed-initiative strategy (MI) (df = 1, F = 28.1, p $<$
.001). Mean Recognition Score for SI was .90, while the Mean
Recognition Score for MI was .72. Furthermore, the performance ranges
were different, with a minimum score for MI of .43, as compared to a
minimum score for SI of .61.  The number of ASR Rejections was also
significantly greater for the MI strategy, with the system rejecting
an average of 1.33 utterances per dialog, as compared with only .44
utterances per dialog for SI (df =1, F = 6.35, p$<$ .02).  However,
despite the poorer ASR performance that we predicted, the average
number of User Turns was significantly {\bf less} for MI (df =1, F=
6.25, p $<$ .02). The average number of User Turns for SI was 21.9
utterances, as compared with a mean of 15.33 for MI.

We had hoped that users of the MI strategy would avail themselves of
the context-specific help to learn the agent's valid vocabulary. While
use of Help Requests was not significant by strategy, more Timeout Prompts were
played for MI (df = 1, F = 62.4, p $<$ .001).  The mean number of
timeouts per dialog for SI was .94, while the mean for MI was 4.2.
Timeout Prompts suggest to the user what to say, and are triggered by
occasions in which the user says nothing after a system utterance,
perhaps because they don't know what they {\bf can} say.  For example,
the most commonly played prompt for MI was {\it You can access
messages using values from the sender or the subject field. If you
need to know a list of senders or subjects, say `List senders', or
`List subjects'. If you want to exit the current folder, say `I'm done
here'.}

In terms of user satisfaction measures, there were no differences in
the Task Ease measure as a function of strategy; users did not think
it was easier to find relevant messages using the SI agent than the MI
agent, even on the first day. Users' perceptions of whether Elvis is
sluggish to respond (System Response) also did not vary as a function
of strategy, probably because the response delays were due to the
application module, which is identical for both strategies. However
users did perceive differences in the Interaction Pace of the
system. Users were more likely to perceive the pace of MI as being
{\it too slow} (df =1, F = 14.01, p $< $.001). One possible source of
this perception is that the SI strategy kept users busy more of the
time, specifying small incremental pieces of information. This would
be consistent with claims about graphical user interfaces
\cite{GTRH95}.  Thus the average pace of interaction in the SI
strategy would be faster, except for those interactions that finally
accessed the specified email folder or message. In contrast, every MI
interaction could result in accessing the email application module, so
on average each interaction was slower paced, despite the fact that
average task completion times were lower.

There was a difference between the MI and the SI agent in users'
perceptions of whether the agent understood them (ASR Performance) (df
=1, F= 14.54, p$<$ .001). This is consistent with the fact that the
Mean Recognition Score was much lower for MI.  Users also perceived
the MI agent as more difficult for them to understand (TTS
Performance) (df =1, F = 4.30, p $<$ .05), possibly because the help
and timeout messages had to be longer for the MI agent, in order to
describe what type of input it could understand.  There is a trend
towards users having more difficulty knowing what to say to the MI
agent (User Expertise) (df =1, F= 3.41, p$<$ .07).

{\bf Task Effects}: Several factors were also significant as a function of task.  As
mentioned above, factors that are significant as a function of task
are potential indicators of learning effects.  Mean Recognition Score
was significant as a function of task (df = 2, F = 4.2, p $<$
.03). The Mean Recognition Score for MI for task 1 was .69, for task 2
was .68, and for task 3 was .80, showing a potential learning effect
of adapting to the system's language limitations over successive
task. Mean recognition score for SI did not improve over task, in fact
showing evidence that task 2 was more difficult, with task 1 mean
recognition at .95, task 2 at .82 and task 3 at .94.

The average number of ASR Rejections per dialog was also significant
as a function of task (df =2, F = 3.3, p$<$ .05). ASR Rejections
averaged 1.33 for MI for task 1, 2.0 for task 2, and .67 for task
3. For SI, ASR rejections averaged .16 for task 1, 1.0 for task 2, and
.16 for task 3.

The number of Help Requests per dialog was significant as a function
of task (df = 2, F = 4.8, p$<$ .03). Users usually asked for help on
the first task but not afterwards.  Users' perceptions of knowing what
they could say (User Expertise) also improved over successive tasks
for both versions of the system (df =2, F = 4.67, p $<$ .02), showing
the largest improvement for MI, as we hypothesized.  For this
question, 1 was mapped to {\it almost never} while 5 was mapped to
{\it almost always}.  The mean for SI was 3.67 for task 1, 2.83 for
task 2 and 4.0 for task 3.  The mean for MI was 2.33 for task 1, 3.00
for task 2, and 3.67 for task 3. Thus at the beginning of the
experiment, most MI subjects thought that they {\it rarely} knew what
to say, and by the end of the experiment, felt that they {\it often}
knew what to say.

{\bf Subject Effects}: Several factors were also significant as a
function of subject.  Some subjects may have had an easier time using
the system.  There were significant differences in Mean Recognition
Score (df = 10, F = 2.7, p $<$ .02), and the frequency with which the
system played Timeout Prompts as a function of subject (df = 10, F=
3.07, p $<$ .01).  We had thought that the use of Barge In might
reflect learning, on the basis that as users acquired more expertise
they would interrupt the system with responses to queries before the
query was completed. However, there was no increase in the number of
Barge Ins over task. There was a significant difference in the use of
Barge In across subjects.  Apparently some subjects felt more
confident about interrupting Elvis.

It is also clear that subjects' perceptions of the system varied.  The
user's perception that they knew what they could say (User Expertise)
differed (df = 10, F = 3.00, p $<$ .01), as well as whether Elvis was
easy to understand (TTS Performance) (df = 10, F = 3.71, p$<$
.005). Perceptions of whether Elvis was slow or sluggish to respond
(System Response) differed (df = 10, F = 2.96, p $<$ .02), as well as
feelings about whether Interaction Pace was appropriate (df = 10, F =
4.84, p $<$ .001). Finally, comparisons of Elvis to the touch tone
interface to voice mail (Comparable Interface) varied across users (df
= 8, F= 3.74, p $<$ .01) .

\section{Performance Function Estimation}

Given this experimental data, we draw on the PARADISE framework to
estimate which factors are most significant in predicting Cumulative
Satisfaction, and thus which factors might form the basis of a
predictive performance function \cite{Walkeretal97}. The overall
structure of objectives in PARADISE that provides the basis for
estimating a performance function is shown in Figure 1.  Cumulative
Satisfaction is the user satisfaction measure in the objectives
structure in Figure 1. The efficiency measures for this experiment are
User Turns, System Turns, and Elapsed Time. The qualitative measures
are Barge Ins, Mean Recognition Score, Timeout Prompts, ASR Rejections
and Help Requests. These qualitative measures reflect the style or the
feel of the interaction.

\begin{figure}[htb]
\centerline{\psfig{figure=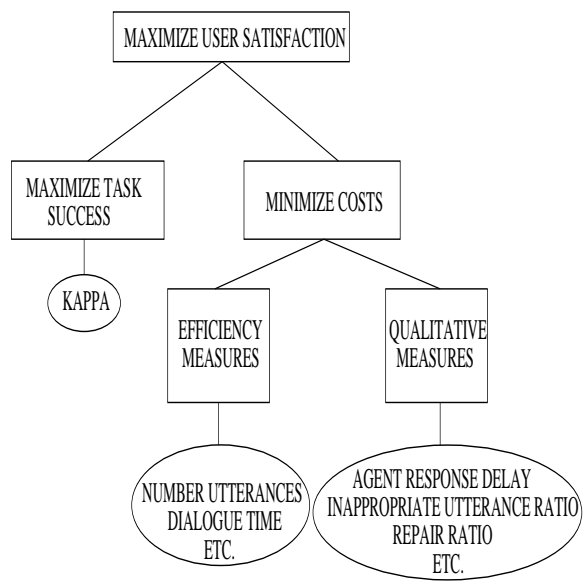,height=3.0in,width=3.0in}}
\caption{PARADISE's structure of objectives for spoken dialog performance}
\label{objectives-fig}
\end{figure}

Multivariate linear regression using all the objective performance
measures shows that the only significant contributors to Cumulative
Satisfaction are User Turns and Mean Recognition Score. The results of
a second regression with only these factors included shows that User
Turns is significant at p $<$ .03, and that Mean Recognition Score is
significant at p$<$ .0001, and that the combination of these two
variables accounts for 42\% of the variance in Cumulative
Satisfaction, the external validation criterion.  Kappa, as a measure
of task success is not a significant variable because subjects
completed  the task in 33 out of 36 cases, leaving very little
variance in the data.  The predicted performance function is :

\[{\rm Performance} = .63 \ast {\cal N}(MeanRecognition) -  .32 \ast {\cal N}( UserTurns)\]

where $\cal N$ is a normalization function that guarantees that the
magnitude of the coefficients is independent of the scales of the
factors.  Applying this performance function to our experimental data,
independent of task, suggests that the SI strategy overall performs
better. The mean performance over all subjects for SI is .214, while
mean performance for MI is -0.213. However, as with the other
measures, the performance of the MI strategy improves over each
successive task, with performance at -0.27 for task 1, rising to 0.125
by task 3.  Continuing the trend that we observe over the three
trials, it seems likely that the performance of MI would outpace that
of SI as users acquire more expertise.

\section{Conclusion}

We discussed the results of an experiment comparing a mixed-initiative
dialog agent with a system-initiative dialog agent, in the context of
a personal agent for accessing email messages by phone. Our initial
hypotheses were that the system-initiative strategy would be better
for inexperienced users, but that as users gained experience with the
system over successive tasks, the mixed-initiative strategy would be
preferred. Our results demonstrated that user's satisfaction and ease
of use with the MI strategy did improve over successive
tasks. However, the overall performance function derived from the
experimental data showed that the MI strategy did not surpass the SI
strategy over the three tasks. Future experiments will give users more
experience to test this hypothesis further.

\section{Acknowledgments}
Bruce Buntschuh, Candace Kamm and Russ Ritenour provided useful help
on questions about using the spoken dialog platform, and to our
subjects for participating in the experiment.

\begin{footnotesize}

\end{footnotesize}
\end{sloppypar}

\end{document}